\documentclass[sigconf, screen, authorversion, nonacm, pagebackref,breaklinks,colorlinks,allcolors=cvprblue]{acmart}
\AtBeginDocument{%
  }

\setcopyright{acmlicensed}
\copyrightyear{2026}
\acmYear{2026}
\acmDOI{XXXXXXX.XXXXXXX}
\acmConference[MM '26]{Proceedings of the 34th ACM International Conference on Multimedia}{October 2026}{TBD}
\acmISBN{978-1-4503-XXXX-X/2018/06}

\usepackage[table]{xcolor}

\definecolor{TabColor1}{RGB}{203,233,157}
\definecolor{TabColor2}{RGB}{153,214,164}
\definecolor{TabColor3}{RGB}{098,190,166}
\definecolor{TabColor4}{RGB}{057,143,185}
\definecolor{TabColor5}{RGB}{078,098,171}

\definecolor{TabColorWarm1}{RGB}{246,163,97}
\definecolor{TabColorWarm2}{RGB}{230,109,80}

\definecolor{cvprblue}{rgb}{0.21,0.49,0.74}
\definecolor{B1}{rgb}{0.21,0.49,0.74}
\definecolor{B2}{rgb}{0.31,0.60,0.85}
\definecolor{B3}{rgb}{0.41,0.70,0.94}
\definecolor{C1}{RGB}{50,70,110}
\definecolor{C2}{RGB}{31,120,180}
\definecolor{C3}{RGB}{106,172,217}
\definecolor{ACMRed}{RGB}{230,100,100}
\usepackage{hyperref}
\usepackage[ruled, linesnumbered]{algorithm2e}
\usepackage{graphicx} 
\usepackage{multirow}
\usepackage{algpseudocode}
\usepackage{multicol}
\usepackage{listings}
\usepackage{amsmath} 
\usepackage{natbib}
\usepackage{wrapfig}
\usepackage{enumitem}
\usepackage[table]{xcolor}

\newcommand{\Fref}[1]{Figure~\ref{#1}}

\definecolor{richpurple}{RGB}{123, 63, 180}
\definecolor{mauve}{RGB}{186, 94, 134}
\definecolor{softcoral}{RGB}{250, 128, 114}

\newcommand{\Yes}{\textcolor{green}{\ding{51}}}
\newcommand{\No}{\textcolor{red}{\ding{55}}}

\usepackage{booktabs}
\usepackage{pifont}
\usepackage{bm}
\usepackage{adjustbox}
\usepackage[most]{tcolorbox}
\usepackage{etoolbox}
\usepackage{cuted}

\tcbuselibrary{breakable} 

\begin{document}

\title{Embodied3DBench: Benchmarking Low-Level Embodied Spatial Intelligence of Vision Language Models}

\author{Jiyao Zhang}
\affiliation{
  \institution{CFCS, School of CS, PKU}
  \city{Beijing}
  \country{China}
}

\author{Mingxu Zhang}
\affiliation{
  \institution{BUPT}
  \city{Beijing}
  \country{China}
}

\author{Yitong Peng}
\affiliation{
  \institution{PKU}
  \city{Beijing}
  \country{China}
}

\author{Haoxuan Liu}
\affiliation{
  \institution{USTB}
  \city{Beijing}
  \country{China}
}

\author{Chenshuo Wang}
\affiliation{
  \institution{PKU}
  \city{Beijing}
  \country{China}
}

\author{Yuxing Long}
\affiliation{
  \institution{CFCS, School of CS, PKU}
  \city{Beijing}
  \country{China}
}

\author{Haoyang Huang}
\affiliation{
  \institution{Jingdong Technology Information Technology Co., Ltd}
  \city{Beijing}
  \country{China}
}

\author{Dongjiang Li}
\affiliation{
  \institution{Jingdong Technology Information Technology Co., Ltd}
  \city{Beijing}
  \country{China}
}

\author{Nan Duan}
\affiliation{
  \institution{Jingdong Technology Information Technology Co., Ltd}
  \city{Beijing}
  \country{China}
}

\author{Hui Shen}
\affiliation{
  \institution{Jingdong Technology Information Technology Co., Ltd}
  \city{Beijing}
  \country{China}
}

\author{Hao Dong}
\authornote{Corresponding author: Hao Dong. Email: hao.dong@pku.edu.cn}
\affiliation{
  \institution{CFCS, School of CS, PKU}
  \city{Beijing}
  \country{China}
}

\begin{abstract}

Are current Vision Language Models (VLMs) ready to comprehend and reason about complex embodied interactions in 3D environments?
We introduce \textbf{Embodied3DBench}, a robot-centric benchmark targeting \textbf{low-level spatial intelligence} in embodied 3D environments.
To systematically evaluate these foundational perceptual capabilities, the benchmark includes \textbf{6} task categories divided into two core groups: Spatial Structural Understanding (Grounding, Spatial Relation Prediction, and Multi-view Correspondence) and Interaction-Oriented Perception (Affordance Prediction, Grasp Point Prediction, and Trajectory Prediction).
The benchmark spans \textbf{12} subcategories and contains over \textbf{21k} high-quality question-answer pairs.
We evaluate 13 state-of-the-art models, and the results show that while current models exhibit relatively strong high-level spatial reasoning, such as understanding object-to-object positional relations, they remain fragile in interaction-oriented perception, highlighting a significant lack of robust \textbf{3D-aware interaction priors}.
To actively bridge this capability gap revealed by our benchmark, we further synthesize a large-scale training dataset comprising \textbf{1.3M} QA pairs.
Notably, fine-tuning on this dataset yields significant improvements in low-level spatial intelligence.
Ultimately, \textbf{Embodied3DBench} fills a critical gap by providing both a systematic evaluation framework and a scalable data solution, setting a clear target for the development of interaction-aware multimodal systems.
\end{abstract}   



\begin{strip}
    \centering
    \vspace{-30pt}
    \includegraphics[width=\textwidth]{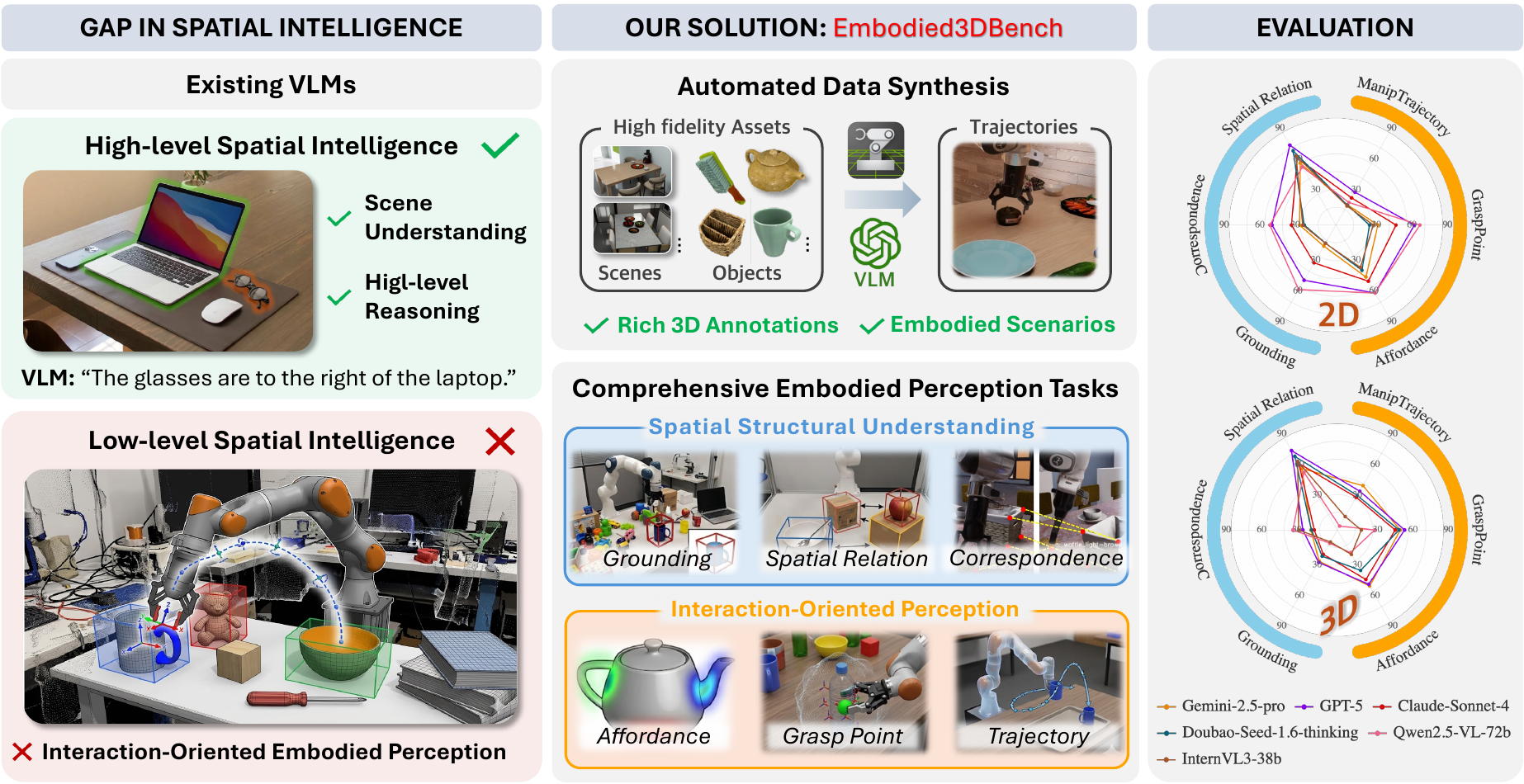}
    \vspace{-20pt}
    \captionof{figure}{Current VLMs exhibit a critical gap between high-level reasoning and low-level embodied spatial understanding. To bridge this gap, we introduce Embodied3DBench, a comprehensive benchmark that systematically evaluates VLMs across six core low-level tasks, categorized into spatial structural understanding and interaction-oriented perception. Extensive evaluations demonstrate that such low-level spatial intelligence remains a significant challenge for state-of-the-art VLMs.}
    \label{fig:teaser}
    \vspace{-10pt}
\end{strip}

\maketitle 

\vspace{-5pt}
\section{Introduction}
\label{sec:intro}

\begin{table*}[htbp]
\centering
\caption{\textbf{Comparison with other Spatial Understanding Benchmarks.} Embodied3DBench addresses low-level embodied spatial intelligence. It is specifically designed to concentrate entirely on embodied perception tasks, featuring a highly diverse set of task categories and a rich abundance of QA pairs.}
\label{tab:compare_exist_benchmark}
\vspace{-8pt}
\adjustbox{max width=1.0\textwidth}
{\begin{tabular}{lcccccl}
\toprule
\textbf{Name} & \textbf{QA} & \textbf{Task Cat.} & \textbf{3D Part-Level Aff.} & \textbf{Multi-View} & \textbf{3D Emb. Manip.} & \textbf{Scenario} \\
\midrule
\midrule
CVBench~\cite{tong2024cambrian1fullyopenvisioncentric} & 3.8K & 4 & \No & \No & \No & General 2D Visual Comprehension \\
BLINK~\cite{fu2024blinkmultimodallargelanguage} & 2.6K & 14 & \No & \Yes & \No & General 2D Visual Comprehension \\
\midrule
RoboSpatial-Home~\cite{Song_2025_CVPR} & 6K & 3 & \No & \No & \No & Indoor, Tabletop \\
EmbSpatial-Bench~\cite{du-etal-2024-embspatial} & 3.6K & 6 & \No & \No & \No & Indoor \\
Spatial-Aptitute-Training~\cite{ray2025satdynamicspatialaptitude} & 150 & 5 & \No & \Yes & \No & Indoor\\
VSI-Bench~\cite{yang2025thinkingspacemultimodallarge} & 5K & 8 & \No & \Yes & \No & Indoor \\
Gemini-Robotics-ERQA~\cite{geminiroboticsteam2025geminiroboticsbringingai} & 400 & 8 & \No & \Yes & \No & Indoor, Robotic Manipulation \\
\midrule
\rowcolor{gray!20}\textbf{Embodied3DBench(Ours)}& 21K & 12 & \Yes & \Yes & \Yes & Tabletop, Robotic Manipulation \\
\bottomrule
\end{tabular}}
\vspace{-10pt}
\end{table*}

Spatial reasoning is a cornerstone of intelligent agents, enabling them to connect visual perception with purposeful embodied interaction~\cite{huang2024rekepspatiotemporalreasoningrelational,11128671,qu2025spatialvlaexploringspatialrepresentations,pan2025omnimanip}. As shown in~\Fref{fig:teaser}, recent advances in large-scale Vision Language Models (VLMs) have demonstrated impressive \emph{high-level spatial intelligence}~\cite{zeng2024lvdiffusor,long2025checkmanual,gao2025realappliance,yu2026correctnav} — the ability to interpret scene layouts, identify coarse object-to-object relations such as ``left of'' or ``in front of'' and reason about semantic context~\cite{Ji_2025_CVPR,baairobobrainteam2025robobrain20technicalreport}. Such capacities have driven progress in domains like autonomous driving~\cite{Chen_2024_CVPR,10.1007/978-3-031-72943-0_15,tian2024drivevlmconvergenceautonomousdriving,10611485}, AR/VR scene understanding~\cite{fan2025vlm3rvisionlanguagemodelsaugmented,mao2025spatiallmtraininglargelanguage,Song_2025_CVPR,Song_2025_CVPR,pan2025metaspatialreinforcing3dspatial}, and embodied navigation~\cite{hirose2025omnivlaomnimodalvisionlanguageactionmodel,zhang2025mem2egoempoweringvisionlanguagemodels,zhang2025nava3understandinginstructionnavigating,zeng2025janusvlndecouplingsemanticsspatiality,zhang2025embodiednavigationfoundationmodel}.

However, transitioning from passive observation to active embodied interaction demands a distinct capability: \emph{low-level spatial intelligence}~\cite{zhang2023generative,zhang2024omni6dpose}. This involves precise multi-view localization, geometry- and topology-aware reasoning in three-dimensional space, and inferring vital physical interaction priors, such as optimal grasp points and actionable spatial trajectories~\cite{li2025controlvlafewshotobjectcentricadaptation,häon2025mechanisticinterpretabilitysteeringvisionlanguageaction,Chen_2024_CVPR,zhangcadgrasp,liu2024rgbgrasp,zhang2026hipolicy}. Unlike high-level reasoning, which suffices for describing object relations, low-level reasoning must integrate fine-grained spatial perception with an understanding of the physical and kinematic constraints of the environment~\cite{zhang2025upvlaunifiedunderstandingprediction,lin2025evo0visionlanguageactionmodelimplicit}.

Consider a tabletop grasping task: knowing that the mug is in front of the kettle may help locate it, but planning a successful interaction depends on estimating the mug’s exact 6D pose, reasoning about its handle position and orientation, and predicting a collision-free spatial trajectory. Similarly, intending to insert a screwdriver into a fixture demands precise geometric alignment and clearance estimation. These scenarios highlight why low-level spatial intelligence is indispensable for bridging ``seeing'' and the perceptual preparation for ``doing'' in embodied agents~\cite{chen2025clutterdexgrasp,wu2023learning}.

Despite its importance, the development of low-level spatial intelligence has progressed slowly, and dedicated benchmarks remain scarce~\cite{guo2025surdsbenchmarkingspatialunderstanding,xu2025deepphybenchmarkingagenticvlms,song2026awaking}, as shown in Table \ref{tab:compare_exist_benchmark}. Two major factors contribute to this gap: (\emph{i}) large-scale, high-quality 3D data from real environments is limited, making it difficult to train models that require strong spatial grounding~\cite{Lu_2025_CVPR,ahmed2025kestrel3dmultimodalllm}; (\emph{ii}) annotating low-level spatial information — such as precise object geometry, pose, affordances, or actionable trajectory priors — is inherently challenging, especially when only RGB images without corresponding 3D measurements are available~\cite{cao2025physx3dphysicalgrounded3dasset}. These constraints have long hindered systematic benchmarking and model development.

\begin{figure*}[htbp]
\centering
\includegraphics[width=0.94\textwidth]{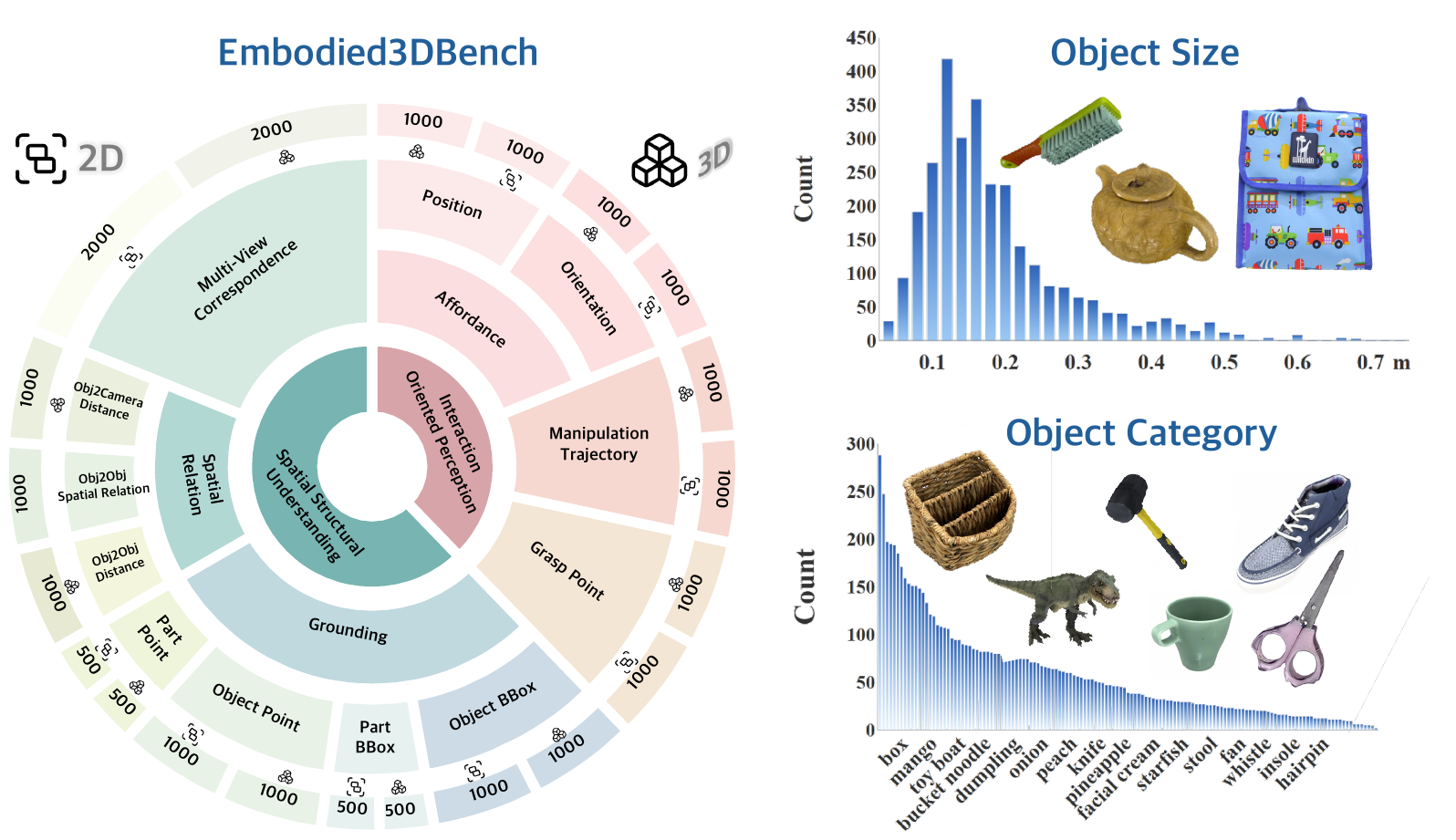}
\vspace{-10pt}
\caption{Embodied3DBench is a large-scale benchmark for low-level embodied spatial intelligence, systematically evaluating models across six task systems divided into two core categories: \emph{Spatial Structural Understanding} (Grounding, Spatial Relation Prediction, and Multi-view Correspondence) and \emph{Interaction-Oriented Perception} (Affordance Prediction, Grasp Point Prediction, and Robot Trajectory Prediction). The objects in the benchmark are shown in the figure on the right, encompass diverse categories and scales, fully covering typical everyday home application scenarios.}
\label{fig:tasks}
\vspace{-10pt}
\end{figure*}

To overcome these challenges, we build \textbf{Embodied3DBench}, as shown in~\Fref{fig:tasks}, a large-scale, robot-centric benchmark explicitly targeting low-level spatial intelligence. Leveraging a controlled high-fidelity simulation data generation pipeline~\cite{dai2022domain}, Embodied3DBench systematically evaluates foundational perceptual capabilities through six task systems divided into two core categories: \emph{Spatial Structural Understanding} (Grounding, Spatial Relation Prediction, and Multi-view Correspondence) and \emph{Interaction-Oriented Perception} (Affordance Prediction, Grasp Point Prediction, and Robot Trajectory Prediction).

We benchmark 13 state-of-the-art VLMs and observe a pronounced dichotomy: models exhibit strong performance on high-level semantic and coarse spatial reasoning tasks, but show fragile competence in inferring low-level, interaction-oriented spatial priors. Models that achieve high accuracy in spatial relation prediction frequently exhibit large error rates in metric-sensitive grounding, grasp point localization, and trajectory prior prediction, highlighting a persistent limitation in current multimodal intelligence. To actively bridge this capability gap revealed by our benchmark, we further synthesize a massive training dataset comprising 1.3M QA pairs, which demonstrates significant effectiveness in enhancing models' low-level spatial intelligence. In summary, our work makes the following contributions:

\begin{itemize}[leftmargin=*]

\item We develop a robot-centric simulation pipeline tailored for \emph{embodied 3D low-level spatial intelligence}, enabling controlled multi-view data generation and fine-grained geometric annotation.
\item We construct \textbf{Embodied3DBench}, a large-scale evaluation framework categorizing tasks into \emph{Spatial Structural Understanding} and \emph{Interaction-Oriented Perception} to systematically assess the pre-execution perceptual capabilities of VLMs.
\item We extensively evaluate 13 state-of-the-art VLMs, identifying a persistent dichotomy between high-level spatial reasoning and the ability to infer robust, 3D-aware interaction priors.
\item We synthesize a massive training dataset comprising \textbf{1.3M} QA pairs and demonstrate that fine-tuning on this data improves the low-level embodied spatial intelligence of the existing model.
\end{itemize}

\section{Related Work}
\label{sec:related}

\subsection{High-Level Spatial Understanding}

Recent progress in visual reasoning benchmarks has primarily focused on evaluating high-level and coarse-grained spatial relationships within 2D scenes. BLINK~\cite{fu2024blinkmultimodallargelanguage} assesses comprehensive perceptual abilities such as multi-view reasoning, depth estimation, and reflectance prediction, covering diverse visual cues. CV-Bench~\cite{tong2024cambrian1fullyopenvisioncentric} divides tasks into 2D and 3D settings: while the 2D tasks evaluate spatial relationship understanding, the 3D tasks are limited to global depth and distance comparison. However, these benchmarks largely rely on internet-scale image datasets, which lack the strict 3D grounding and kinematic context required for inferring embodied interaction priors.

Notably, state-of-the-art VLMs demonstrate strong proficiency in handling these advanced spatial comprehension tasks—specialized models like SpatialVLM~\cite{Chen_2024_CVPR} and SpatialRGPT~\cite{NEURIPS2024_f38cb4cf} achieve reliable performance on benchmarks for coarse-grained spatial reasoning, including spatial relationship judgment and orientation recognition. Even general-purpose foundational models like GPT-4V~\cite{yang2023dawnlmmspreliminaryexplorations} and Gemini~\cite{comanici2025gemini25pushingfrontier} have competitive accuracy across the advanced reasoning dimensions covered by these benchmarks. Their performance validates that the core capabilities of high-level spatial understanding have been effectively mastered by current large models.

\subsection{Embodied Spatial Reasoning}

To bridge the gap between static internet images and dynamic physical environments, recent embodied benchmarks have extended evaluation to 3D spaces, covering indoor navigation and tabletop interaction scenarios. SAT~\cite{ray2025satdynamicspatialaptitude} introduced systematic dynamic spatial reasoning by considering both self-motion (e.g., camera rotation) and object motion, generating low-cost and high-quality QA pairs through 3D simulation. RoboSpatial~\cite{Song_2025_CVPR} focuses on reasoning about free space around objects—determining whether a given object can fit into specific empty regions—thereby improving spatial awareness for potential object placement. Other benchmarks, such as EmbSpatial-Bench~\cite{du-etal-2024-embspatial}, emphasize egocentric spatial reasoning by assessing large vision-language models (LVLMs) in first-person embodied environments. Furthermore, RoboBrain-2.0~\cite{baairobobrainteam2025robobrain20technicalreport} has demonstrated that mixed training on spatial datasets, coupled with Chain-of-Thought (CoT) and Reasoning-via-Feedback (RFT), significantly stimulates models' reasoning capabilities. These works collectively demonstrate that enhanced 3D spatial awareness benefits downstream embodied planning. However, they still operate primarily at the scene- or object-level of abstraction, neglecting the fine-grained, geometry-critical structural understanding that is indispensable for reasoning about actionable interaction.

\subsection{Low-Level Perception}

Fine-grained low-level perception has been shown to facilitate physical interaction by providing detailed structural cues and interaction-oriented priors. A3VLM~\cite{huang2024a3vlmactionablearticulationawarevision} predicts bounding boxes and joint axes from purely 2D observations, providing crucial geometric parameters for planning motion primitives such as sliding and rotational actions. Sofar~\cite{qi2025sofarlanguagegroundedorientationbridges} learns semantic vector representations of objects to bridge the gap between abstract task-level descriptions and actionable pose estimation. Although recent methods have leveraged part-level inference to enhance downstream execution, the systematic evaluation of large models’ native capabilities in low-level geometric and interaction reasoning remains largely unexplored~\cite{wang2025omniearbenchmarkingagentreasoning,Mo_2019_CVPR,Geng_2023_CVPR,han2025robocerebralargescalebenchmarklonghorizon}. Existing literature lacks an established benchmark that quantitatively assesses both fine-grained geometric localization and the prediction of kinematic priors. This necessitates the development of a comprehensive benchmark—like Embodied3DBench—that explicitly evaluates the continuum from \emph{spatial structural understanding} to \emph{interaction-oriented perception}, ensuring multimodal models possess the robust pre-execution intelligence required for real-world environments.

\section{Embodied3DBench}
\label{sec:method}

\subsection{Overview}
\label{sec:overview}

In order to systematically evaluate the low-level spatial intelligence of Vision-Language Models (VLMs), we propose Embodied3DBench. We define 12 sub-tasks that are inherently challenging to resolve by relying solely on coarse-grained visual semantics. Each task requires geometry-aware reasoning, tightly coupled with embodied interaction requirements. These span from foundational structural comprehension (e.g., part-level grounding) to advanced interaction-oriented reasoning (e.g., predicting affordances).

\vspace{-5pt}
\subsection{Task Definition}
\label{sec:task_definition}

We introduce 6 tasks to evaluate models’ low-level spatial intelligence, grouped into \textbf{Spatial Structural Understanding} and \textbf{Interaction-Oriented Perception}. Each task features 2D and 3D variants, and the 3D version requires metric 3D predictions.

\noindent\textbf{Spatial Structural Understanding Tasks}

\begin{itemize}[leftmargin=*]

    \item \textbf{Grounding.}
        This task comprises two distinct sub-tasks: pointing and bounding box prediction. 
    
        \textbf{Pointing}, 
        utilizing input 2D images to precisely locate and mark target objects or spatial part-level positions specified by the user through semantic descriptions. 
            
        \textit{Example: `Locate the silver handle of the pot.'}
        
        \textbf{Bounding Box}, 
        requiring the target object to be fully enclosed within an axis-aligned bounding box. The model must infer the box’s center coordinates, dimensions, and rotation.
    
        \textit{Example: `Locate the inner wall of the light gray bowl in the image and provide the 3D bounding box results.'}
    
    \item \textbf{Spatial Relation.}
        Focusing on the relative positional and orientational relationships between entities in 3D space. This task is divided into two subtasks: \textbf{relative distance estimation} and \textbf{relative direction determination}.
    
        \textit{Example:} \item[1)]\leftskip=1.5em 
            \textit{`Where is the book in relation to the toy? (A) Left \space\space (B) Right'}
            \item[2)]\leftskip=1.5em 
            \textit{`What is the distance from the battery to the camera in meters?'}

    \item \leftskip=0em \textbf{Multi-View Correspondence.}
        Identifies cross-viewpoint feature correspondences. Given a specific point’s coordinates from viewpoint A, the model must determine whether it is visible from viewpoint B; if so, output its precise coordinates in B, otherwise return `not visible.'
        
        \textit{Example: `The yellow body of the banana is located at (673, 363) in the first image. Please confirm if this specific point is visible in the second image; if so, provide its coordinates.'}

\end{itemize}

\noindent\textbf{Interaction-Oriented Perception Tasks}

\begin{itemize}[leftmargin=*]

    \item \textbf{Affordance Prediction.} This dimension focuses on the essential perceptual prerequisite of `how objects can be interacted with' ~\cite{4456755}. It predicts the functional attributes and regions required for task execution, divided into 2 subcategories: 
        
        \textbf{Functional Point}, which corresponds to the key physical location where an embodied agent must directly exert an effect during interaction. 
        
        \textbf{Functional Vector}, which infers the spatial direction of motion required to complete an effective interaction.

        \textit{Example: `You need to pull the drawer out. Please provide its functional point and vector.'}
    
    \item \textbf{Trajectory Prediction.} The prediction of a continuous spatial motion path that an end-effector would follow during interactions with the environment, serving as a vital kinematic prior.
    
        \textit{Example: `Pick the watch and place it on the dressing table. Please predict the key trajectory points needed to complete this task from this viewpoint onward.'}

    \item \textbf{Grasp Point Prediction.}
        Predicting the optimal region on an object for a robot to reliably establish control, ensuring sufficient grasping force and postural stability prior to execution.

        \textit{Example: `Where is the optimal grasp point of the toy bus?'}
    
\end{itemize}

\begin{figure*}[t]
    \centering
    \includegraphics[width=\textwidth]{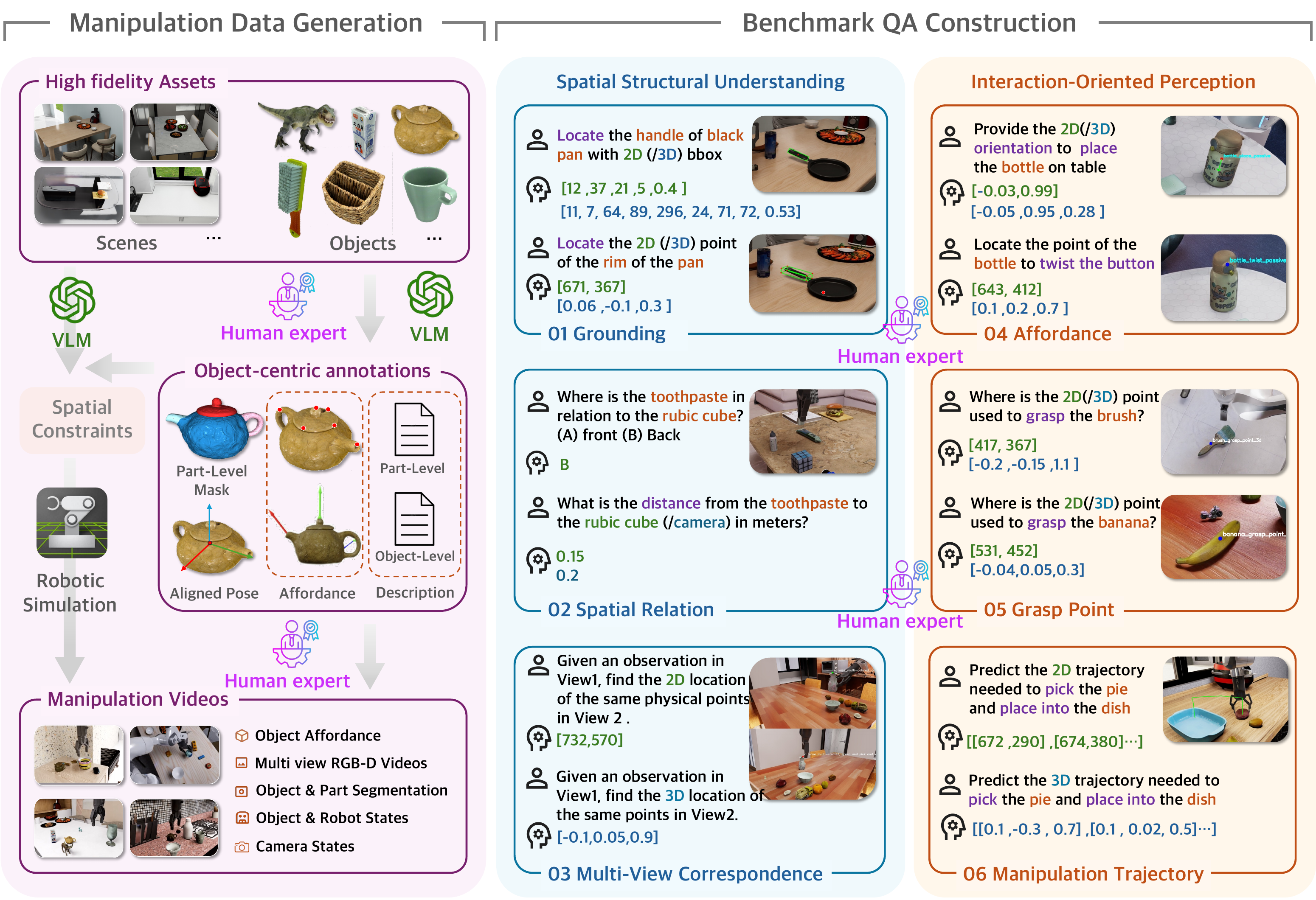}
    \vspace{-20pt}
    \caption{\textbf{Embodied3DBench Construction Pipeline.} High-fidelity interaction data synthesis is achieved through a fully automated pipeline encompassing asset annotation, kinematic constraint definition, and physics-based rendering, with human quality checks at critical stages. This high-quality embodied data is used to sample appropriate visual frames, while task-specific geometric rules are applied to extract ground-truth spatial priors and generate QA pairs via a template-based approach. Finally, rigorous manual quality inspection ensures data reliability.}
    \vspace{-10pt}
    \label{fig:benchmark_construction}
\end{figure*}

\vspace{-10pt}
\subsection{Benchmark Construction}
\label{sec:benchmark_construction}

\noindent\textbf{Embodied Interaction Data Synthesis.}
Our goal is to synthesize high-quality, large-scale embodied data in the simulation to serve as the foundation for low-level QA generation, as illustrated in Figure \ref{fig:benchmark_construction}. The process is formulated as follows:
\begin{itemize}[leftmargin=*]
    \item \textbf{Asset Collection}: Gather high-quality 3D scene and object scan data from sources like Omni6DPose~\cite{10.1007/978-3-031-73226-3_12}. Perform object-centric annotations using VLMs (e.g., interaction modes, component-level semantic segmentation). Conduct multiple rounds of manual quality inspection.
    \item \textbf{Task Definition}: VLMs divide simulated tasks into sub-phases based on the required object interactions, define spatial constraints, and establish kinematic parameters~\cite{hua2024gensim2scalingrobotdata,chen2025robotwin20scalabledata}.
    \item \textbf{Hi-Fi Data Generation}: Render interaction trajectories using Isaac Sim according to task sub-phases. Output per-frame RGB-D images, component-level segmentation maps, and precise 6D pose data (camera, objects, and agent).
    \item \textbf{Embodied Data Quality Inspection}: Validate data across dimensions including visual occlusion, task completion logic, and trajectory plausibility. Filter out invalid episodes to construct a high-fidelity embodied simulation dataset.
\end{itemize}

\noindent\textbf{QA Pair Generation.}
Leveraging the synthesized embodied videos and annotated 3D assets, we extract fine-grained spatial priors. To mitigate ambiguity caused by uneven object visibility, we apply two strict filtering rules: `Invisibility Exclusion' (removing frames outside the FOV using position thresholds, $p \le 0$ or $p \ge W$) and `Occlusion Exclusion' (filtering heavily occluded frames using pixel-area, $area \le 3000px^{2}$, and an occlusion ratio criteria, $occluded\_ratio \ge 0.3$). This ensures that subsequent QA queries remain geometrically unambiguous. Visible frames are then grouped by task and processed via coordinate transformations, spatial feature extraction, and projection.

\begin{itemize}[leftmargin=*]
    \item \textbf{Affordance.} Transform the 3D functional point and interaction vector from object coordinates to the camera space. For 2D variants, both are projected onto the camera plane to obtain the 2D interaction direction.
    \item \textbf{Object and Part Pointing.} Using a coordinate transformation logic similar to Affordance, the 3D target point is converted into a 2D precise positioning coordinate within the image plane.
    \item \textbf{Bounding Box.} In 3D, the oriented bounding box is calculated via coordinate transformation in camera space. In 2D, the bounding box is computed from the object or part segmentation mask, recording its center, dimensions, and rotation.
    \item \textbf{Grasp Point.} Based on the `close gripper' command moment recorded in the physics simulation, we extract the end-effector's 6D pose information at that moment as the GT grasp point.
    \item \textbf{Multi-View Correspondence.} Employ Farthest Point Sampling~\cite{NIPS2017_d8bf84be} to select three candidate points per object part. For each part–frame–view quadruple, visibility is recorded in a lookup table; during generation, these entries are cross-referenced to extract coordinates across visible views.
    \item \textbf{Trajectory.} Perform uniform temporal sampling of the end-effector's motion trajectory. Project these 3D trajectory waypoints onto the camera coordinate system, generating both 2D projection and 3D metric trajectory priors simultaneously.
    \item \textbf{Spatial Relation.} Direction is defined along three orthogonal axes: up/down, left/right, and front/back. The up/down and left/right directions are derived from projected pixel coordinates. The front/back direction is strictly labeled only when the objects' 3D bounding boxes do not overlap along any axis, utilizing their average depth values for relative depth determination. 
\end{itemize}

\noindent\textbf{Human-in-the-Loop Verification.} Initial QA construction revealed several edge cases, including limited viewpoint diversity, visual ambiguity from partial occlusion, and mismatches between semantic annotations and 3D geometry. We developed a three-step hierarchical visual inspection platform to mitigate these issues and guarantee data reliability:
\begin{itemize}[leftmargin=*]
    \item For each 3D asset, human annotators verify the accuracy of semantic labels and functional descriptions, correcting any geometric discrepancies.
    \item Based on scene diversity and task validity metrics, we strictly filter video samples from the massive dataset, ensuring comprehensive scene coverage without physics or rendering errors.
    \item The generated QA data is presented via an intuitive visual interface. Human reviewers examine each question–answer pair to verify its absolute geometric alignment with the visual inputs, ensuring unambiguous spatial grounding.
\end{itemize}

\noindent\textbf{Benchmark Statistics \& Dataset Scaling.} Through random combinations of 4.8K object instances and large-scale scene assets, we systematically generated \textbf{21K} high-quality, fine-grained QA pairs, forming the rigorous evaluation benchmark shown in Figure ~\ref{fig:tasks}. Furthermore, to actively bridge the capability gap revealed in our evaluation, we scaled up this generation pipeline to synthesize a massive, dedicated training collection comprising \textbf{1.3M} QA pairs. 

\subsection{View-Augmented Chain-of-Thought}
\label{sec:va_cot}
Chain-of-Thought (CoT) prompting has proven effective in complex reasoning by guiding models to articulate intermediate steps before reaching a final conclusion. This insight prompts a critical question: can we design a CoT paradigm specifically tailored for 3D spatial perception, where geometric relations, occlusions, and viewpoint transformations play a central role? To explore this, we introduce View-Augmented Chain-of-Thought (\textbf{VA-CoT}), which extends conventional CoT by integrating supplementary novel views of the scene. These extra perspectives serve as explicit spatial cues, enabling the model to mentally align object geometry across viewpoints, reason effectively about hidden or partially occluded regions, and consolidate robust 3D structural relationships during its step-by-step reasoning process. This provides a structured framework to study how multi-view visual evidence can enhance foundational spatial intelligence for embodied tasks.

\begin{table*}[htbp]
    \centering
    \caption{\textbf{Evaluation results of state-of-the-art VLMs on Embodied3DBench.} We systematically evaluate models across the dual spectrum of spatial structural understanding and interaction-oriented perception. Results indicate that both proprietary and open-source models, including high-performing systems like GPT-5 and the Gemini series, still struggle significantly with fine-grained, geometry-critical spatial intelligence.}
    \label{tab:results_six_ability}
    \adjustbox{max width=1.0\textwidth}{\newcolumntype{E}{>{\centering\arraybackslash}p{0.8cm}}
\newcolumntype{F}{>{\centering\arraybackslash}p{1cm}}
\newcolumntype{G}{>{\centering\arraybackslash}p{1.8cm}}
\newcolumntype{D}{>{\centering\arraybackslash}p{2.4cm}}

\begin{tabular}{lG|EEFFD|FFGGEE|EE}
\toprule
\multirow{3}{*}{\textbf{Model}} 
  & \multirow{3}{*}{\textbf{Reasoning}}
  & \multicolumn{5}{c|}{\textbf{Spatial Structural Understanding}} 
  & \multicolumn{6}{c|}{\textbf{Interaction-Oriented Perception}}
  & \multirow{2}{*}{\textbf{Avg.}} \\
\cmidrule(lr){3-13}
  &
  & \multicolumn{2}{c}{\textbf{Grounding}} 
  & \multicolumn{2}{c}{\textbf{Correspondence}} 
  & \multirow{1}{*}{\textbf{Spatial Relation}}
  & \multicolumn{2}{c}{\textbf{Affordance}}
  & \multicolumn{2}{c}{\textbf{Manipulation Trajectory}} 
  & \multicolumn{2}{c|}{\textbf{Grasp Point}} \\
\cmidrule(lr){3-15}
  & 
  & 2D & 3D 
  & 2D & 3D 
  &
  & 2D & 3D
  & 2D & 3D
  & 2D & 3D 
  & 2D & 3D\\
\midrule

\rowcolor{gray!20} \multicolumn{15}{l}{Proprietary Models} \\
\midrule

Gemini-2.5-pro~\cite{team2023gemini} & \Yes & 20.5 & \cellcolor{ACMRed!60}30.7 & 30.7 & 22.4 & 60.2 & 50.6 & \cellcolor{ACMRed}54.8 & 19.6 & \cellcolor{ACMRed!60}43.4 & 35.4 & \cellcolor{ACMRed!20}53.3 & 31.4 & \cellcolor{ACMRed!60}44.1 \\
Gemini-2.5-flash~\cite{team2023gemini} & \Yes & 21.8 & 26.3 & 32.2 & 22.6 & 67.4 & 50.3 & 43.0 & 20.1 & 36.9 & 34.4 & \cellcolor{ACMRed!60}53.5 & 31.8 & \cellcolor{ACMRed!20}41.6 \\
GPT-5~\cite{openai} & \Yes & \cellcolor{ACMRed!60}53.7 & \cellcolor{ACMRed}31.7 & \cellcolor{ACMRed!20}53.9 & 29.2 & \cellcolor{ACMRed}77.7 & \cellcolor{ACMRed!60}66.1 & \cellcolor{ACMRed!60}53.1 & \cellcolor{ACMRed!60}32.0 & \cellcolor{ACMRed!20}38.0 & \cellcolor{ACMRed!20}66.2 & \cellcolor{ACMRed}57.0 & \cellcolor{ACMRed!20}54.4 & \cellcolor{ACMRed}47.8 \\
GPT-4o~\cite{hurst2024gpt} & \No & 28.7 & 17.1 & 30.5 & 36.0 & 63.3 & 53.0 & 30.7 & 24.6 & 11.8 & 46.2 & 29.0 & 36.6 & 31.3 \\
Claude-Sonnet-4 & \No & 36.8 & 23.8 & 37.5 & 19.5 & 65.2 & 54.6 & \cellcolor{ACMRed!20}48.4 & \cellcolor{ACMRed!20}26.4 & 32.0 & 50.7 & 49.0 & 41.2 & 39.7 \\
Doubao-Seed-1.6-thinking & \Yes & 17.6 & 25.7 & 26.7 & 21.9 & \cellcolor{ACMRed!60}72.2 & 44.0 & 39.5 & 19.3 & 33.8 & 28.3 & 52.2 & 27.2 & 40.9 \\
Doubao-Seed-1.6 & \No & 15.8 & \cellcolor{ACMRed!20}30.0 & 27.3 & 29.8 & 63.7 & 38.2 & 41.5 & 20.7 & 29.9 & 27.0 & 51.7 & 25.8 & 41.1 \\
Doubao-Seed-1.5-thinking-vision-pro~\cite{guo2025seed1} & \Yes & 16.9 & 23.5 & 28.4 & 21.3 & \cellcolor{ACMRed!20}67.8 & 41.8 & 34.8 & 22.6 & 34.7 & 27.6 & 47.4 & 27.4 & 38.3 \\

\midrule
\rowcolor{gray!20} \multicolumn{15}{l}{Open-Source Models} \\
\midrule

Qwen2.5-VL-72b~\cite{wang2024qwen2} & \No & \cellcolor{ACMRed}62.8 & 12.9 & \cellcolor{ACMRed!60}55.7 & \cellcolor{ACMRed!60}37.7 & 56.5 & \cellcolor{ACMRed}66.1 & 22.2 & 21.9 & 3.8 & \cellcolor{ACMRed!60}70.8 & 31.6 & \cellcolor{ACMRed!60}55.5 & 27.5 \\
Qwen2.5-VL-7b~\cite{wang2024qwen2} & \No & 48.5 & 5.3 & 36.2 & \cellcolor{ACMRed!20}37.2 & 59.8 & \cellcolor{ACMRed!20}55.0 & 20.0 & 26.3 & 5.0 & 49.1 & 12.6 & 43.0 & 23.3 \\
InternVL3.5-8b~\cite{wang2025internvl3} & \No & 19.9 & 10.4 & 31.0 & 29.3 & 60.4 & 37.9 & 25.1 & 22.0 & 6.7 & 33.9 & 19.6 & 28.9 & 25.2 \\
InternVL3-38b~\cite{zhu2025internvl3exploringadvancedtraining} & \No & 17.8 & 11.7 & 27.9 & 32.7 & 67.3 & 41.2 & 23.9 & 18.6 & 13.2 & 31.5 & 20.3 & 27.4 & 28.2 \\
Qwen3-VL-4B~\cite{yang2025qwen3technicalreport} & \No & \cellcolor{ACMRed!20}49.6 & 21.0 & \cellcolor{ACMRed}58.3 & \cellcolor{ACMRed}42.1 & 65.2 & 51.9 & 28.5 & \cellcolor{ACMRed}37.6 & \cellcolor{ACMRed}44.5 & \cellcolor{ACMRed}84.9 & 39.4 & \cellcolor{ACMRed}56.5 & 40.1 \\
\bottomrule

\midrule
\rowcolor{gray!20} \multicolumn{15}{l}{Human Evaluation} \\
\midrule

Human &  & \textbf{85.3} & \textbf{46.6} & \textbf{79.2} & \textbf{66.1} & \textbf{72.7} & \textbf{69.4} & \textbf{42.7} & \textbf{55.5} & \textbf{68.6} & \textbf{84.8} & \textbf{70.2} & \textbf{74.8} & \textbf{61.2} \\

\bottomrule
\end{tabular}
}
\end{table*}

\section{Experiments}
\label{sec:experiments}

\subsection{Experimental Settings}
\label{subsec:experimental_settings}

We test all models under a zero-shot protocol. Our benchmark encompasses leading VLMs from diverse families, including both proprietary and open-source systems, operating in standard inference and advanced reasoning modes. Proprietary models are accessed via official APIs, while open-source models are deployed locally under their standard configurations to ensure strict reproducibility.

\subsection{Evaluation Metrics}
\label{sec:eval-metrics}

To ensure a fair and rigorous evaluation, we design specific metrics tailored to the varying output modalities. We categorize the evaluation metrics based on the required geometric output format:

\begin{itemize}[leftmargin=*]
    \item \textbf{Bounding Box.}
    We employ the \textbf{Intersection-over-Union (IoU)} (utilizing the 3D BBox estimator from~\cite{mousavian20173d}) as the metric. 
    
    \item \textbf{Coordinate Point.}
    We utilize the $\ell_2$ distance as the metric. For 2D tasks, this is measured in pixel-level; for 3D tasks, it is evaluated as the absolute Euclidean distance measured in meters.

    \item \textbf{Directional Vector.}
    Cosine similarity is employed as the metric.
    
    \item \textbf{Spatial Trajectory.}
    We adopt the \textbf{Hausdorff Distance}~\cite{chen2011clustering} to rigorously quantify the topological and metric dissimilarity. 
    
    \item \textbf{Categorical Choice.}
    We directly report the classification accuracy of the predicted spatial relations.

    \item \textbf{Metric Distance.}
    We use the \textbf{absolute error} between the predicted distance and the ground-truth distance. All values are evaluated in Euclidean space and measured in meters. 
    
\end{itemize}

To unify the disparate ranges of these metrics, we apply appropriate mapping functions tailored to the value domain of each raw metric. Specifically, we utilize two non-linear mappings, $x^{\alpha}$ and $e^{-\alpha x}$, to obtain the normalized score $s \in [0, 100]$. The detailed formulation and hyperparameter $\alpha$ settings for each task are provided in the supplementary material.

\begin{table}[htbp]
    \centering
    \caption{\textbf{Grounding Evaluation Results.} This experiment isolates the models' capabilities in fine-grained spatial structural understanding, specifically evaluating absolute metric localization at both the object and part levels.}
    \label{tab:results_grounding}
    \adjustbox{max width=0.48\textwidth}{\begin{tabular}{l*{2}{c}*{2}{c}*{2}{c}*{2}{c}}
\toprule
\multirow{2}{*}{\textbf{Model}} & \multicolumn{2}{c}{\textbf{Object BBox}} & \multicolumn{2}{c}{\textbf{Part BBox}} & \multicolumn{2}{c}{\textbf{Object Point}} & \multicolumn{2}{c}{\textbf{Part Point}} \\
\cmidrule(lr){2-9}
   & 2D & 3D & 2D & 3D & 2D & 3D & 2D & 3D \\
\midrule
\rowcolor{gray!20} \multicolumn{9}{l}{Proprietary models} \\
\midrule
Gemini-2.5-pro~\cite{team2023gemini} & 9.1 & 6.8 & 4.9 & \cellcolor{ACMRed!20}3.2 & 34.0 & \cellcolor{ACMRed!60}55.9 & 31.5 & \cellcolor{ACMRed!60}55.3 \\
Gemini-2.5-flash~\cite{team2023gemini} & 7.2 & 3.8 & 5.9 & 2.5 & 37.1 & 49.7 & 36.4 & 48.7 \\
GPT-5~\cite{openai} & \cellcolor{ACMRed!60}45.4 & \cellcolor{ACMRed}8.7 & \cellcolor{ACMRed!60}30.3 & \cellcolor{ACMRed}5.6 & 68.3 & \cellcolor{ACMRed}57.0 & 64.7 & \cellcolor{ACMRed!20}53.4 \\
GPT-4o~\cite{hurst2024gpt} & 13.3 & 1.1 & 9.9 & 0.1 & 46.4 & 35.5 & 42.7 & 29.2 \\
Claude-Sonnet-4 & 21.6 & 3.6 & 16.6 & 2.2 & 53.9 & 43.9 & 53.2 & 45.6 \\
Doubao-Seed-1.6-thinking & 4.4 & 6.4 & 3.2 & 2.7 & 31.5 & 46.5 & 30.7 & 45.9 \\
Doubao-Seed-1.6 & 3.2 & \cellcolor{ACMRed!60}8.3 & 2.3 & \cellcolor{ACMRed!60}4.0 & 29.0 & \cellcolor{ACMRed!20}53.3 & 28.5 & 52.7 \\
Doubao-Seed-1.5-thinking-vision-pro~\cite{guo2025seed1} & 3.3 & \cellcolor{ACMRed!20}7.4 & 2.4 & 3.2 & 31.4 & 42.4 & 29.5 & 38.2 \\
\midrule
\rowcolor{gray!20} \multicolumn{9}{l}{Open-Source models} \\
\midrule
Qwen2.5-VL-72b~\cite{wang2024qwen2} & \cellcolor{ACMRed}49.0 & 0.2 & \cellcolor{ACMRed}40.5 & 0.1 & \cellcolor{ACMRed}82.3 & 23.0 & \cellcolor{ACMRed!60}73.8 & 31.2 \\
Qwen2.5-VL-7b~\cite{wang2024qwen2} & \cellcolor{ACMRed!20}24.1 & 0.1 & 15.8 & 0.0 & \cellcolor{ACMRed!20}77.4 & 7.7 & \cellcolor{ACMRed!20}72.0 & 16.4 \\
InternVL3.5-8b~\cite{wang2025internvl3} & 3.5 & 0.2 & 4.6 & 0.0 & 36.3 & 19.6 & 35.0 & 22.8 \\
InternVL3-38b~\cite{zhu2025internvl3exploringadvancedtraining} & 2.5 & 0.7 & 2.4 & 0.1 & 32.1 & 24.4 & 35.0 & 19.8 \\
Qwen3-VL-4B~\cite{yang2025qwen3technicalreport} & 21.9 & 3.0 & \cellcolor{ACMRed!20}19.2 & 0.6 & \cellcolor{ACMRed!60}80.3 & 30.2 & \cellcolor{ACMRed}74.1 & \cellcolor{ACMRed}58.9 \\
\bottomrule
\end{tabular}}
\end{table}

\begin{table}[htbp]
    \centering
    \caption{\textbf{Spatial Relation Evaluation Results.} This assessment probes the models' structural reasoning, evaluating the ability to correctly determine metric and topological relationships both `between objects' and `between objects and the camera' in 3D space.}
    \label{tab:results_spatial_relation}
    \adjustbox{max width=0.48\textwidth}{\newcolumntype{E}{>{\centering\arraybackslash}p{2.8cm}}
\newcolumntype{D}{>{\centering\arraybackslash}p{3.2cm}}

\begin{tabular}{lEDE}
\toprule
\multirow{2}{*}{\textbf{Model}} & \multicolumn{1}{c}{\textbf{Obj2Cam Distance}} & \multicolumn{1}{c}{\textbf{Obj2Obj Spatial Rel.}} & \multicolumn{1}{c}{\textbf{Obj2Obj Distance}} \\
\cmidrule(lr){2-4}
   & 3D & 3D & 3D \\
\midrule
\rowcolor{gray!20} \multicolumn{4}{l}{Proprietary models} \\
\midrule
Gemini-2.5-pro~\cite{team2023gemini} & 63.4 & 42.9 & 74.4 \\
Gemini-2.5-flash~\cite{team2023gemini} & 63.4 & 67.5 & 71.2 \\
GPT-5~\cite{openai} & \cellcolor{ACMRed}67.1 & \cellcolor{ACMRed}88.7 & \cellcolor{ACMRed!60}77.2 \\
GPT-4o~\cite{hurst2024gpt} & 52.4 & 66.3 & 71.3 \\
Claude-Sonnet-4 & 50.8 & 76.9 & 67.9 \\
Doubao-Seed-1.6-thinking & 58.8 & \cellcolor{ACMRed!60}82.0 & 75.9 \\
Doubao-Seed-1.6 & 58.8 & 58.8 & 73.6 \\
Doubao-Seed-1.5-thinking-vision-pro~\cite{guo2025seed1} & \cellcolor{ACMRed!20}63.6 & 61.4 & \cellcolor{ACMRed}78.4 \\
\midrule
\rowcolor{gray!20} \multicolumn{4}{l}{Open-Source models} \\
\midrule
Qwen2.5-VL-72b~\cite{wang2024qwen2} & 49.2 & 47.3 & 72.9 \\
Qwen2.5-VL-7b~\cite{wang2024qwen2} & 51.5 & 64.2 & 63.7 \\
InternVL3.5-8b~\cite{wang2025internvl3} & 59.3 & 55.0 & 66.8 \\
InternVL3-38b~\cite{zhu2025internvl3exploringadvancedtraining} & 51.7 & \cellcolor{ACMRed!20}79.1 & 71.1 \\
Qwen3-VL-4B~\cite{yang2025qwen3technicalreport} & \cellcolor{ACMRed!60}63.9 & 55.7 & \cellcolor{ACMRed!20}76.0 \\
\bottomrule
\end{tabular}}
\end{table}

\begin{table}[htbp]
    \centering
    \caption{\textbf{Affordance Evaluation Results.} This experiment evaluates the crucial interaction-oriented perception capabilities, testing whether the functional point of contact and the actionable vector can be accurately inferred prior to execution.}
    \label{tab:results_affordance}
    \adjustbox{max width=0.48\textwidth}{\begin{tabular}{l*{2}{c}*{2}{c}}
\toprule
\multirow{2}{*}{\textbf{Model}} & \multicolumn{2}{c}{\textbf{Position}} & \multicolumn{2}{c}{\textbf{Orientation}} \\
\cmidrule(lr){2-5}
   & 2D & 3D & 2D & 3D \\
\midrule
\rowcolor{gray!20} \multicolumn{5}{l}{Proprietary models} \\
\midrule
Gemini-2.5-pro~\cite{team2023gemini} & 30.6 & \cellcolor{ACMRed}51.5 & \cellcolor{ACMRed}70.6 & \cellcolor{ACMRed}58.1 \\
Gemini-2.5-flash~\cite{team2023gemini} & 36.4 & 47.3 & 64.1 & 38.7 \\
GPT-5~\cite{openai} & \cellcolor{ACMRed!20}64.3 & \cellcolor{ACMRed!20}51.2 & \cellcolor{ACMRed!60}67.8 & \cellcolor{ACMRed!60}55.0 \\
GPT-4o~\cite{hurst2024gpt} & 41.6 & 24.9 & \cellcolor{ACMRed!20}64.4 & 36.5 \\
Claude-Sonnet-4 & 46.2 & 44.0 & 63.1 & \cellcolor{ACMRed!20}52.9 \\
Doubao-Seed-1.6-thinking & 30.9 & 40.7 & 57.1 & 38.4 \\
Doubao-Seed-1.6 & 27.2 & \cellcolor{ACMRed!60}51.3 & 49.3 & 31.8 \\
Doubao-Seed-1.5-thinking-vision-pro~\cite{guo2025seed1} & 29.2 & 33.7 & 54.3 & 36.0 \\
\midrule
\rowcolor{gray!20} \multicolumn{5}{l}{Open-Source models} \\
\midrule
Qwen2.5-VL-72b~\cite{wang2024qwen2} & \cellcolor{ACMRed}71.1 & 11.7 & 61.1 & 32.7 \\
Qwen2.5-VL-7b~\cite{wang2024qwen2} & \cellcolor{ACMRed!60}64.9 & 7.6 & 45.2 & 32.3 \\
InternVL3.5-8b~\cite{wang2025internvl3} & 27.6 & 14.4 & 48.1 & 35.8 \\
InternVL3-38b~\cite{zhu2025internvl3exploringadvancedtraining} & 28.2 & 15.0 & 54.1 & 32.8 \\
Qwen3-VL-4B~\cite{yang2025qwen3technicalreport} & 50.3 & 31.7 & 53.5 & 25.2 \\
\bottomrule
\end{tabular}}
    \vspace{-20pt}
\end{table}

\subsection{Main Results}
\label{subsec:main_results}

Table~\ref{tab:results_six_ability} reports the comprehensive model performance across the Embodied3DBench pipeline. Tables~\ref{tab:results_grounding}, \ref{tab:results_spatial_relation}, and \ref{tab:results_affordance} provide granular metric breakdowns for Grounding, Spatial Relation, and Affordance sub-tasks, respectively. Our extensive evaluation reveals three critical insights:

\noindent\textbf{1. Current VLMs lack robustness in precise geometric encoding.}
Across all tested model families, performance degrades sharply in tasks demanding absolute metric localization rather than coarse semantic description. For instance, in 3D \textit{Object BBox} and \textit{Object Point} tasks, even GPT-5—the most capable proprietary reasoning model—yields only mid-range accuracy. This exposes a fundamental dichotomy: while models excel at high-level spatial narratives, their internal representations lack the strict geometric constraints required for embodied structural understanding.

\noindent\textbf{2. Performance is asymmetric across perception paradigms.}
While a few elite models dominate the overall leaderboard, their proficiency is unevenly distributed. GPT-5 achieves the highest global average, driven by strong zero-shot reasoning in multi-view correspondence and affordance tasks. Gemini-2.5-Pro demonstrates competitive capability in geometry-grounded interaction tasks, such as inferring 3D affordance vectors and actionable trajectories. Interestingly, certain open-source systems exhibit highly specialized strengths: Qwen3-VL-4B not only outperforms all proprietary models in 2D perception but also demonstrates highly competitive performance in 3D perception. We attribute this success to the extensive integration of spatial perception tasks during its pre-training phase.

\noindent\textbf{3. Interaction-oriented perception remains the primary bottleneck.}
The benchmark clearly indicates that inferring kinematic priors (grasp points, trajectory planning) is significantly harder for current VLMs than static structural understanding. No single system currently offers uniformly robust capabilities across accurate metric grounding and pre-execution interaction prediction, exposing a critical vulnerability in existing VLMs.

\subsection{In-Depth Analysis}
\label{subsec:in_depth_analysis}

\begin{figure}[ht]
    \centering
    \includegraphics[width=0.36\textwidth]{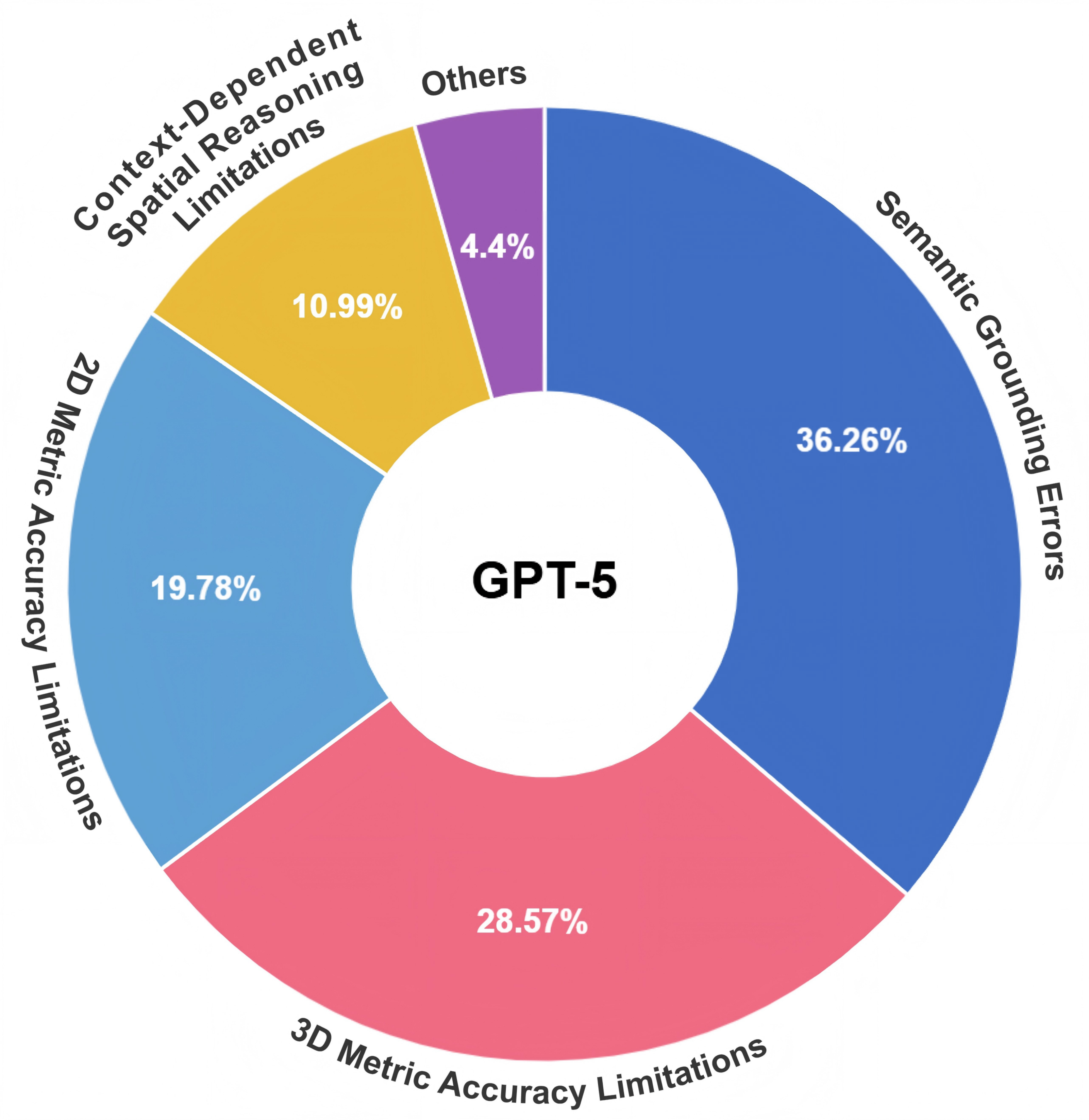}
    \vspace{-8pt}
    \caption{\textbf{Error Type Breakdown of GPT-5.}}
    \label{fig:gpt5_error_breakdown}
    \vspace{-8pt}
\end{figure}

Given GPT-5's consistently leading performance across multimodal perception tasks, we select it as the representative benchmark for an in-depth mechanistic analysis. We isolate two fundamental dimensions—perceptual grounding and metric spatial accuracy—where the capability boundaries of modern VLMs are most distinctly revealed. Figure~\ref{fig:gpt5_error_breakdown} visualizes the taxonomy of GPT-5's errors, providing quantitative backing for our analysis.

\noindent\textbf{The Semantic-to-Geometric Binding Gap.}
As shown in Figure~\ref{fig:gpt5_error_breakdown}, \textit{Semantic Grounding Errors} constitute the largest proportion (36.3\%). These represent scenarios where GPT-5 correctly comprehends the global scene context but fails to accurately bind specific semantic instructions to precise geometric features (e.g., misidentifying functional sub-parts). This indicates that while macroscopic semantic alignment is mature, the fine-grained mapping from "semantic symbols" to "local visual geometries" frequently breaks down. Furthermore, \textit{Context-Dependent Spatial Reasoning Limitations} account for 11.0\% of failures, occurring when correct grounding depends heavily on relative topological cues. This suggests that GPT-5 struggles to dynamically update its spatial graph representation when multiple relational constraints are introduced.

\noindent\textbf{Deficiencies in Metric Spatial Accuracy.}
Errors directly related to coordinate precision represent the most critical hurdle for interaction-oriented perception. \textit{2D Metric Accuracy Limitations} make up 19.8\% of the errors, typically manifesting as bounding boxes that correctly capture object scale but are spatially offset in the image plane. 
Crucially, predicting exact coordinates in 3D space proves substantially more difficult: \textit{3D Metric Accuracy Limitations} dominate with 28.6\% of total errors. Unlike 2D projection, 3D inference demands the implicit understanding of camera intrinsics and depth scale. The high error rate here corroborates that metric-level 3D reasoning—the bedrock of actionable interaction priors—remains an unsolved challenge for vision-language architectures trained primarily on internet-scale 2D images.

\begin{table}[htbp]
    \centering
    \caption{\textbf{Impact of VA-CoT on GPT-5's 3D Spatial Intelligence.} The integration of View-Augmented CoT consistently enhances the model's structural understanding and interaction prior prediction capabilities across challenging 3D sub-tasks.}
    \vspace{-5pt}
    \label{tab:cot_results}
    \adjustbox{max width=0.48\textwidth}
    {\begin{tabular}{lccc}
\toprule
\textbf{Task} & \textbf{GPT-5} & \textbf{GPT-5-CoT} & \textbf{Improvement} \\
\midrule

Object BBox(3D) & 8.7 & 8.9 & {\textbf{+0.2} ~$\uparrow$} \\
Part BBox(3D) & 5.6 & 8.3 & {\textbf{+2.7} ~$\uparrow$} \\
Object Point(3D) & 57.0 & 59.7 & {\textbf{+2.7}~$\uparrow$} \\
Grasp Point(3D) & 57.0 & 57.6 & {\textbf{+0.6}~$\uparrow$} \\
Affordance Position(3D) & 51.2 & 52.3 & {\textbf{+1.1}~$\uparrow$} \\
Affordance Orientation(3D) & 55.0 & 56.9 & {\textbf{+1.9}~$\uparrow$} \\

\bottomrule
\end{tabular}}
    \vspace{-15pt}
\end{table}

\noindent\textbf{Effect of VA-CoT on 3D Reasoning.}
To investigate whether explicit spatial prompting can mitigate these metric limitations, we evaluate the impact of View-Augmented Chain-of-Thought (VA-CoT) on 3D spatial tasks. As demonstrated in Table~\ref{tab:cot_results}, equipping GPT-5 with VA-CoT—providing supplementary viewpoints to enforce multi-view consistency—yields uniform improvements. The most significant gains are observed in fine-grained structural localization tasks (\textit{Part BBox}: +2.7, \textit{Object Point}: +2.7). This confirms that supplying structured, cross-view geometric evidence explicitly aids the model in mentally aligning semantic cues with precise 3D coordinates. We also observe moderate enhancements in interaction-oriented tasks (\textit{Affordance Position}: +1.1, \textit{Affordance Orientation}: +1.9), indicating that multi-view spatial awareness directly translates into more robust kinematic prior predictions.

\begin{table}[htbp]
    \centering
    \caption{\textbf{Effectiveness of Large-Scale Fine-Tuning.}}
    \vspace{-10pt}
    \label{tab:results_finetune}
    \adjustbox{max width=0.48\textwidth}{\begin{tabular}{lccc}
\toprule
\textbf{Task} & \textbf{Qwen3-VL-4B} & \textbf{Qwen3-VL-4B-finetune} & \textbf{Improvement} \\
\midrule
\rowcolor{gray!20} \multicolumn{4}{l}{2D} \\
\midrule
Grounding & 49.6 & 78.1 & {\textbf{+28.5}~$\uparrow$} \\
Correspondence & 58.3 & 59.4 & {\textbf{+1.1}~$\uparrow$} \\
Affordance & 51.9 & 64.2 & {\textbf{+12.3}~$\uparrow$} \\
Manipulation Trajectory & 37.6 & 54.2 & {\textbf{+16.6}~$\uparrow$} \\
Grasp Point & 84.9 & 89.0 & {\textbf{+4.1}~$\uparrow$} \\
\midrule
Avg. & 56.5 & 69.0 & {\textbf{+12.5}~$\uparrow$} \\
\midrule
\rowcolor{gray!20} \multicolumn{4}{l}{3D} \\
\midrule
Grounding & 21.0 & 61.5 & {\textbf{+40.5}~$\uparrow$} \\
Correspondence & 42.1 & 68.4 & {\textbf{+26.3}~$\uparrow$} \\
Spatial Relation & 65.2 & 78.1 & {\textbf{+12.9}~$\uparrow$} \\
Affordance & 28.5 & 72.1 & {\textbf{+43.6}~$\uparrow$} \\
Manipulation Trajectory & 44.5 & 65.5 & {\textbf{+21.0}~$\uparrow$} \\
Grasp Point & 39.4 & 89.3 & {\textbf{+49.9}~$\uparrow$} \\
\midrule
Avg. & 40.1 & 72.5 & {\textbf{+32.4}~$\uparrow$} \\
\bottomrule
\end{tabular}}
    \vspace{-5pt}
\end{table}

\noindent\textbf{Bridging the Gap: The Impact of Large-Scale Fine-Tuning.}
The zero-shot evaluation exposes a clear deficiency in the native low-level spatial intelligence of existing models. To validate whether high-quality, task-specific data can actively bridge this gap, we utilize our synthesized dataset comprising 1.3M embodied QA pairs. As presented in Table~\ref{tab:results_finetune}, fine-tuning a representative open-source architecture (e.g., Qwen3-VL-4B) on this massive corpus yields profound improvements. The model transitions from fragile, coarse-grained estimation to exhibiting robust metric accuracy and reliable interaction prior prediction. This substantial performance leap not only validates the rigorous design of the Embodied3DBench tasks but also confirms that our automated generation pipeline provides a highly scalable and effective data solution for training the next generation of action-ready, spatially-aware multimodal foundations.

\begin{figure}[ht]
    \centering
    \includegraphics[width=0.5\textwidth]{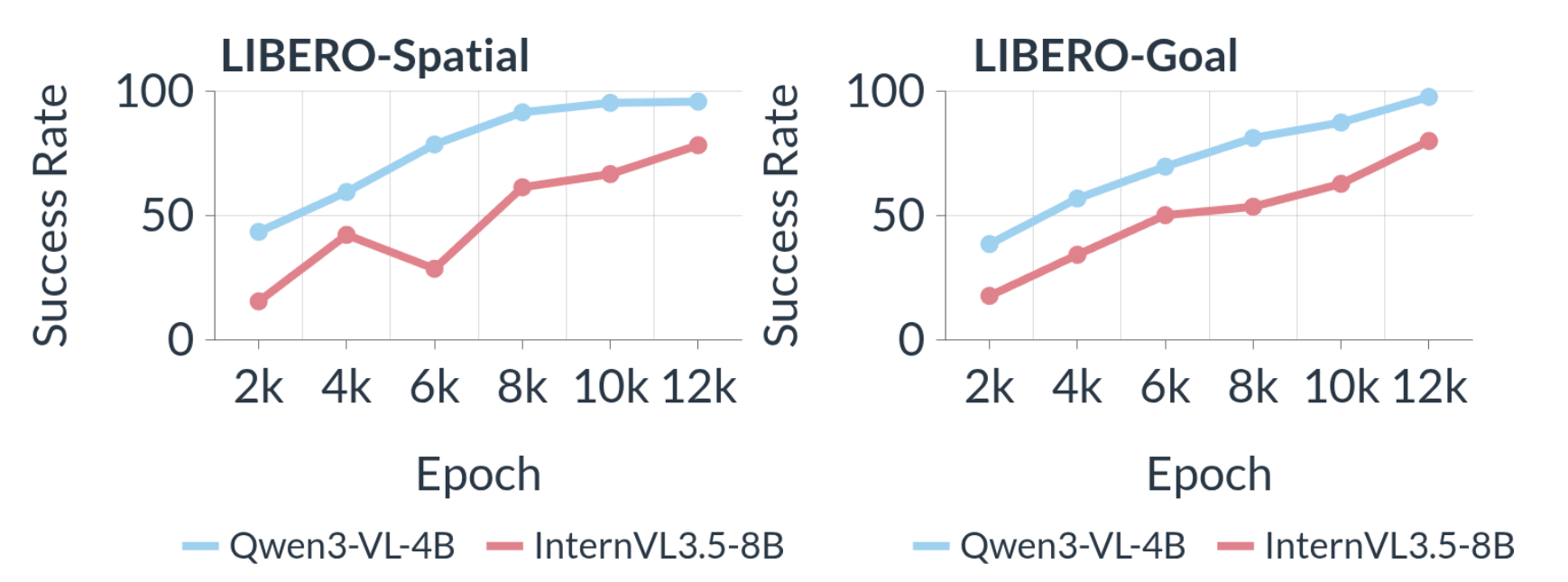}
    \vspace{-20pt}
    \caption{\textbf{Results on LIBERO Benchmark.}}
    \label{fig:libero_results}
    \vspace{-5pt}
\end{figure}

\subsection{Connection to Downstream Tasks.}
To briefly explore whether the capabilities measured by Embodied3DBench reflect a model's utility in downstream embodied tasks, we conduct a preliminary evaluation on the LIBERO benchmark. We train a Vision-Language-Action (VLA) policy by substituting the VLM backbone of $\pi_{0.5}$ with InternVL3.5-8B and Qwen3-VL-4B. As shown in~\Fref{fig:libero_results}, Qwen3-VL-4B, which achieved higher overall scores in our spatial structural and interaction-oriented perception tasks, demonstrates both \textbf{faster convergence} and \textbf{higher final success rates} across the LIBERO-Spatial and LIBERO-Goal suites. This alignment suggests that the foundational spatial priors evaluated in our benchmark can serve as a meaningful indicator of a model's potential effectiveness and learning efficiency in downstream VLA applications.
\section{Conclusion}
\label{sec:conclusion}
We present \textbf{Embodied3DBench} to systematically evaluate MLLMs on \emph{spatial structural understanding} and \emph{interaction-oriented perception}, alongside a 1.3M QA training dataset. Evaluations of 13 state-of-the-art models reveal that while high-level spatial reasoning is robust, inferring metric-sensitive 3D geometric and interaction priors remains a critical bottleneck. Together, our benchmark and dataset offer a scalable foundation for advancing interaction-aware multimodal intelligence.

\noindent\textbf{Limitations and Future Work.}
Currently, the correlation between benchmark performance and downstream utility is validated solely through post-training on the LIBERO suite. Future work will explore leveraging our massive 1.3M dataset for large-scale Vision-Language-Action (VLA) pre-training, aiming to fundamentally inject robust 3D spatial priors directly into foundational models.

{
    \small
    \bibliographystyle{ieeenat_fullname}
    \bibliography{main}
}
\end{document}